\documentclass[conference]{IEEEtran}
\IEEEoverridecommandlockouts
\usepackage{cite}
\usepackage{amsmath,amssymb,amsfonts}
\usepackage{algorithmic}
\usepackage{graphicx}
\usepackage{textcomp}
\usepackage[table,xcdraw]{xcolor}
\usepackage{marvosym}
\usepackage{amsmath,epsfig}
\usepackage{amssymb}
\usepackage{amsmath}
\usepackage{framed,multirow}
\usepackage{latexsym}
\usepackage{indentfirst}
\usepackage{url}
\usepackage{dsfont}
\usepackage{tabularray}
\usepackage{booktabs}
\usepackage{mathrsfs}
\usepackage[ruled,linesnumbered]{algorithm2e}
\usepackage{array}\newcolumntype{C}[1]{>{\centering\arraybackslash}p{#1}}
\usepackage{hyperref}
\usepackage{graphicx}
\usepackage{authblk}

\def\BibTeX{{\rm B\kern-.05em{\sc i\kern-.025em b}\kern-.08em
    T\kern-.1667em\lower.7ex\hbox{E}\kern-.125emX}}
\begin{document}

\title{Self-adaptive vision-language model for 3D segmentation of pulmonary artery and vein

\thanks{
  $\star$ These authors have contributed equally to this work.}
}

\author[2]{\textbf{Xiaotong Guo$^\star$}}
\author[1,3]{\textbf{Deqian Yang$^\star$}}
\author[3]{\textbf{Dan Wang}}
\author[5]{\textbf{Haochen Zhao}}
\author[8]{\textbf{Yuan Li}}
\author[2]{\authorcr {\textbf{Zhilin Sui}}}
\author[4]{\textbf{Tao Zhou}}
\author[1]{\textbf{Lijun Zhang\dag}}
\author[6,7]{\textbf{Yanda Meng\dag}}

\affil[1]{\emph{Key Laboratory of System Software (Chinese Academy of Sciences) and} \authorcr {State Key Laboratory of Computer Science, Institute of Software, Chinese Academy of Sciences, China.}}
\affil[2]{\emph{Department of Thoracic Surgery, National Clinical Research Center for Cancer/Cancer Hospital Shenzhen Hospital}}
\affil[3]{School of Intelligent Science and Technology, Hangzhou Institute for Advanced Study,
\authorcr {University of Chinese Academy of Sciences, Hangzhou 310024, China.}}
\affil[4]{R\&D Center, Guangxi Huayi Artificial Intelligence Medical Technology Co., Ltd}
\affil[5]{School of Computer Science and Engineering, Beihang University, Beijing, China.}
\affil[6]{Department of Computer Science, University of Exeter, Exeter, UK.}
\affil[7]{Department of Cardiovascular \& Metabolic Medicine, University of Liverpool, Liverpool, UK.}
\affil[8]{Guangzhou Jiayi Software Technology Co., Ltd.}
\affil[ \dag ]{\emph{Corresponding author:} 
\emph{zhanglj@ios.ac.cn , Y.m.meng@exeter.ac.uk }}


\maketitle

\begin{abstract}
Accurate segmentation of pulmonary structures is crucial in clinical diagnosis, disease study, and treatment planning. Significant progress has been made in deep learning-based segmentation techniques, but most require much labeled data for training. Consequently, developing precise segmentation methods that demand fewer labeled datasets is paramount in medical image analysis. The emergence of pre-trained vision-language foundation models, such as CLIP, recently opened the door for universal computer vision tasks. Exploiting the generalization ability of these pre-trained foundation models on downstream tasks, such as segmentation, leads to unexpected performance with a relatively small amount of labeled data.
However, exploring these models for pulmonary artery-vein segmentation is still limited. This paper proposes a novel framework called Language-guided self-adaptive Cross-Attention Fusion Framework. Our method adopts pre-trained CLIP as a strong feature extractor for generating the segmentation of 3D CT scans, while adaptively aggregating the cross-modality of text and image representations. We propose a s
pecially designed adapter module to fine-tune pre-trained CLIP with a self-adaptive learning strategy to effectively fuse the two modalities of embeddings. We extensively validate our method on a local dataset, which is the largest pulmonary artery-vein CT dataset to date and consists of 718 labeled data in total. The experiments show that our method outperformed other state-of-the-art methods by a large margin. Our data and code  will be made publicly available upon acceptance.
\end{abstract}


\section{Introduction}
In recent years, pulmonary vascular diseases, including pulmonary embolism and pulmonary hypertension, have emerged as one of the conditions with the highest morbidity and mortality rates. Computed tomography (CT) has been widely adopted as a diagnostic tool to elucidate tomographic patterns of pulmonary diseases \cite{yuan2023pulmonary}. 
Therefore, implementing automated pulmonary vascular segmentation is of significant clinical importance for achieving a three-dimensional reconstruction of the pulmonary vascular architectures. However, the manual delineation process remains labour-intensive due to the complexity of tubular structures. 
Segmentation methods for lung vessels have primarily focused on Convolutional Neural Networks (CNNs), particularly the U-Net architecture and its variants. These approaches have effectively maximized the potential of limited labeled data, especially from CT scans. Many semi-supervised and weakly supervised learning approaches are proposed based on pseudo labeling of the partially labeled data \cite{meng2022dual,meng2023weakly,meng2024multi,meng2022shape,meng2023coarse,meng2022uncertainty,zhao2022focal,zhao2024guidednet,prcvydq}. However, they often suffer significantly from the incorrectness of pseudo labels associated with unlabeled parts of the CT data.

Recently, a new learning paradigm known as the Vision-Language Model (VLM) pre-training and zero-shot prediction has gained significant attention. This paradigm involves pre-training a vision-language model using large-scale image-text pairs abundantly available online.
The pre-trained VLM can be directly applied to downstream visual recognition tasks without fine-tuning. For example, CLIP \cite{radford2021clip} employs an image-text contrastive objective by bringing paired images and texts closer together in the embedding space while pushing unpaired ones further apart. 
Numerous efforts are being made to adapt VLMs to specific task domains. For example, some approaches \cite{radford2021clip,he2020momentum,yang2022unified} modify the contrastive objectives to generative or alignment objectives to retrain a VLM. On the other hand, other methods fine-tune existing VLMs at a lower cost, including techniques such as prompt tuning \cite{liu2023pre} and feature adapters \cite{gao2024clip}.  Regarding medical image segmentation, models such as SAM \cite{kirillov2023segment} and its variants \cite{wang2023sammed3d,ma2024segment} have been retrained on medical images (e.g., Med-SAM \cite{ma2024segment}, SAM-MED3D \cite{wang2023sammed3d}), achieving significant success in this area. However, the required use of box or point prompts is unsuitable for artery-vein segmentation. This is because the ends of pulmonary vessels are quite intricate to locate in CT scans, as illustrated in Fig.~\ref{fig:comperation_results}. This complexity necessitates extensive box or point labeling to achieve satisfactory masks, rendering the process labour-intensive.

This work introduces an efficient language-guided self-adaptive cross-attention fusion framework that integrates adaptive modules designed explicitly for pulmonary artery/vein (A/V) segmentation tasks. Our model not only preserves the performance of the pre-trained model to a great extent but also leverages the unique characteristics of data collected in local hospital settings more effectively. By incorporating these adaptive modules, our framework achieved an average DSC score of 76.22\%  on A/V segmentation, significantly surpassing the performance of the current state-of-the-art methods, such as nnU-Net \cite{isensee2021nnunet} by an average DSC score of 9.03\%, and nnFormer \cite{zhou2021nnformer} by an average DSC score of 12.37\%.
The primary contributions of this study are as follows:

(1) We exploit a large pre-trained vision-language model to segment pulmonary arteries and veins using a substantial local dataset comprising 718 annotated CT scans. The proposed dataset will be made publicly available upon acceptance.

(2) We propose an adaptive module incorporating attention mechanisms and data augmentation methods that are specially designed for our dataset to highlight the vascular characteristics of pulmonary arteries and veins. These mechanisms significantly enhance the fusion of features between the language and visual models, yielding awe-inspiring results on the test dataset.

(3) Extensive experiments on a large number of annotated CT data have been conducted to validate the effectiveness of our proposed methods. The results of these experiments have demonstrated significant performance improvement over other state-of-the-art methods.
\begin{figure*}[!htp]
    \centering
    \includegraphics[scale=0.70]{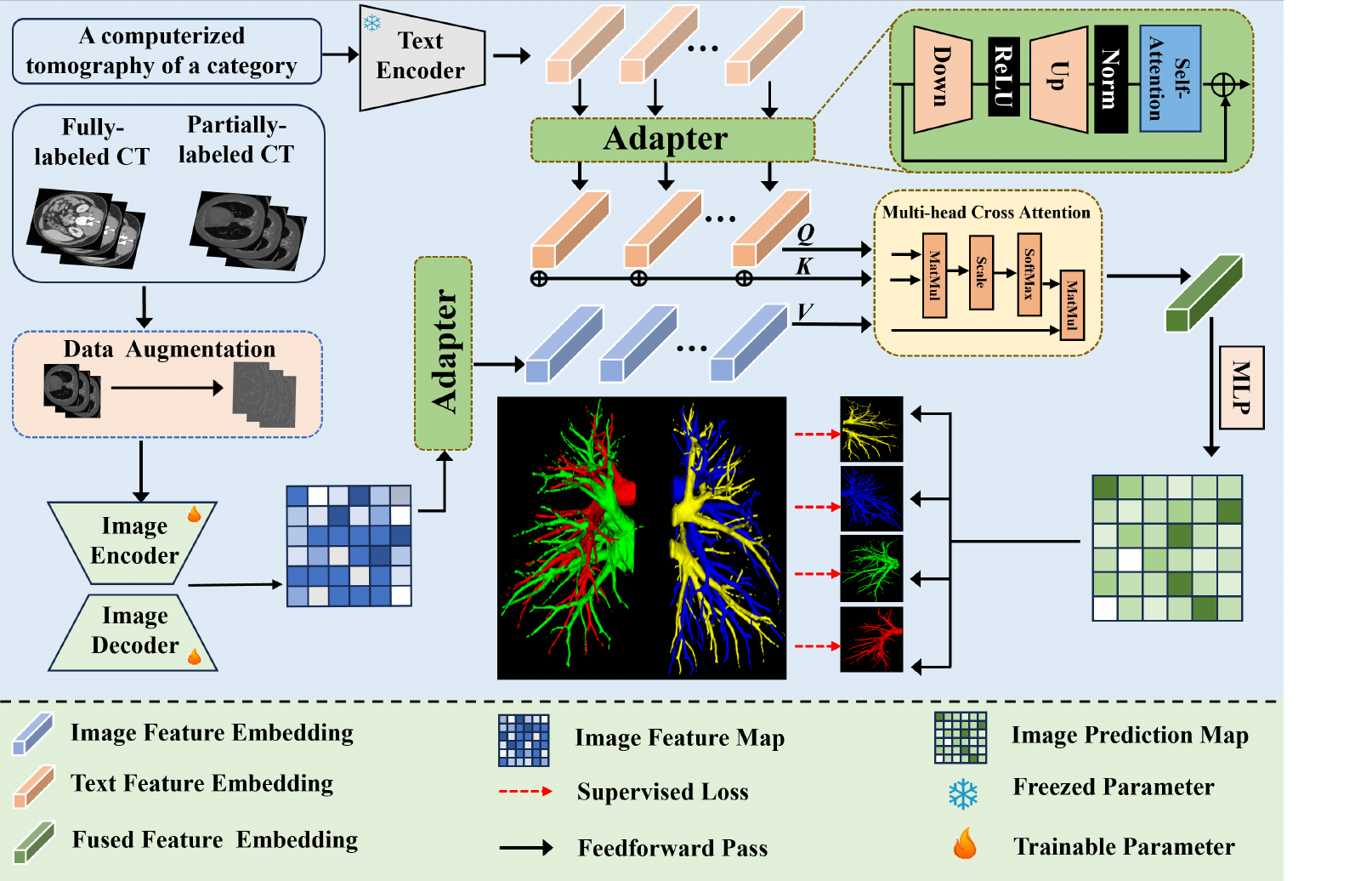}
    
    \caption{
     Overview of the proposed Language-guided self-Adaptive Cross-Attention Fusion Framework, which comprises a text encoder and an image segmentation model. Our model can adaptively learn suitable  embedding for the left/right vein and artery. Best viewed in color. 
    }
    \label{fig:method}
    \end{figure*}

\section{Related Work}

\subsection{Vision language models in medical imaging}

Recently, VLMs have achieved significant advancements in cross-modal tasks and visual recognition problems, as exemplified by models such as CLIP \cite{radford2021clip} and ALIGN \cite{jia2021_align}. CLIP, a pioneering work in large-scale vision-language pre-training, utilized contrastive learning to learn image representations from a dataset of 400 million (image, text) pairs. Inspired by CLIP and ALIGN, a substantial number of medical image-text pairs have been employed to train VLMs through contrastive learning as foundation models. For instance, PMC-CLIP \cite{lin2023pmc} was trained on over 1M medical image-text pairs, while BiomedCLIP \cite{zhang2023biomedclip} was trained on over 15M pairs. 
The majority of their research efforts have been directed towards medical visual question-answering tasks, aiming to provide pre-trained models that capture the similarity between text and medical images. There has been a scarcity of work focusing on the medical image segmentation domain. Notably, Liu et al. \cite{liu2023clip,  zhang2023continual} utilized the original CLIP encoder embedding fused with a segmentation network encoder embedding to address the issue of partial labeling in abdominal organ segmentation, achieving the top rank in the Medical Segmentation Decathlon competition \cite{antonelli2022medical_decathlon}.

\subsection{Pulmonary vascular segmentation}
The segmentation of pulmonary arteries and veins remains an open challenge. In recent years, many studies have adopted 2D or 3D U-Net \cite{qin2021learning_3dunet,lou2024detail_2dunet} to automatically extract features, achieving satisfying performance. However, accurate vascular segmentation remains challenging due to the scarcity of high-quality open-sourced datasets: the extreme complexity of the vascular tree structure, the close proximity (often interwoven) of arteries and veins, and their similar intensity values (especially in non-contrast CT). The Parse2022 \cite{luo2023parse} competition provided 100 CT images for arterial segmentation, and the top-ranked participant used Res-U-Net \cite{diakogiannis2020resunet} combined with threshold segmentation for arterial segmentation. Similarly, Qi et al. \cite{qi2024deep} proposed to extract semantic embedding in a dual U-Net architecture. This study is based on 57 CT datasets from LUNA16 \cite{setio2017validation_luna}, with initial artery and vein labels generated using region growing in the lung vessel mask. Differently, in this work, we propose the largest pulmonary artery-vein dataset to date, which comprises 718 CT scans that are finely annotated by clinicians.

\section{Methods}
CLIP (Contrastive Language-Image Pre-training) is a pretraining method developed by OpenAI \cite{radford2021clip}. Built upon the methodology of contrastive pre-training \cite{liu2021self}, it jointly optimizes a vision encoder and a text encoder, where the vision encoder is based on either ResNet \cite{he2016deep} or Vision Transformer(ViT) \cite{dosovitskiy2020image}. 
The language encoder is rooted in a transformer-based model like BERT \cite{devlin2018bert}, forcing the paired image-text information to be as close as possible to the joint image-text latent space after encoding. We adopt the original CLIP model as our text embedding extractor.  Trained on a vast collection of image-text pairs, CLIP learns visual representation through text supervision, known as prompt. We design a specialized prompt for our pulmonary vessel segmentation task, as seen in Table.~\ref{tabel:clip_prompt}.

\subsection{Pretrained text encoder and vision model}

\textbf{Text encoder:} We use the original pre-trained CLIP encoder $E_{text}$ with a specially designed medical prompt ( \textit{i.e.} `A computerized tomography of a category with small branches') to generate text embeddings $ H_t \in \mathbb{R}^{K * D }$, where K represents the number of class, and D represents the length of the embedding. The pre-trained encoder consists of a 12-layer 512-wide transformer with eight attention heads. The 512-wide output of the transformer is used as text embedding. To enhance the CLIP architecture's medical capability for medical image segmentation tasks, we use K text adapters $A_{text}$ to fine-tune $E_{text}$.  
We observe that the selection of medical prompt templates is hand-crafted and worthy of experiments. Table \ref{tabel:clip_prompt} illustrates the effectiveness of four different prompt templates. The last template, specifically designed for our vascular-shaped data, demonstrates nearly a  0.1\% improvement compared to other commonly used templates, indicating that adjusting the prompt benefits our model.

\begin{table}[!htp]
\centering
\caption{Ablation studys of different prompts.}
\label{tabel:clip_prompt}
\begin{tabular}{ 
    m{1.7cm}<{\centering} | 
    m{4.5cm}<{\centering} | 
    m{1.1cm}<{\centering} 
}

\toprule[2pt]
\scriptsize
Embedding & prompt & \multicolumn{1}{c}{DSC(\%) $\uparrow$ }   \\ 
\midrule[1pt] 
\multicolumn{3}{c}{train:val:test = 502:72:143(Fully labeled + Half-labeled)} \\ 
\midrule[1pt] 
CLIP v1 & A photo of a category &$74.35_{0.36}$ \\ 
\midrule[1pt] 
CLIP v2 & A computerized tomography of a category &$75.74_{0.24}$  \\ 
\midrule[1pt] 
CLIP v3 & There is a category in this computerized tomography &$76.03_{0.37}$ \\ 
\midrule[1pt] 
CLIP v4 & A computerized tomography of a category with small branches &$76.22_{0.48}$ \\ 

\bottomrule[2pt]
\end{tabular}
\end{table}

\begin{align}
H_t &= E_{text}(x_{text}),\\
H_t^a&= A_{text}(H_t).
\end{align}

\textbf{Vision model:}
Medical Segmentation Decathlon~\cite{antonelli2022medical_decathlon} is a benchmark for many medical organ segmentation tasks. Specifically, Liu et al.~\cite{liu2023clip} ranked first with 
an open-source pre-trained model \footnote{https://github.com/ljwztc/CLIP-Driven-Universal-Model} U-Net and Swin UNETR.
Accounting for its strong ability to segment organs, the pre-trained model minimizes the time cost of training a  model and inherits the weights that are suitable for  organ segmentation. Therefore, we adopt a pre-trained U-Net model as the backbone for segmentation. Specifically, in our model, the 3D CT images $x_{img} \in \mathbb{R}^{H*W*L}$ are encoded into a feature map $V \in \mathbb{R}^{B*C*H*W*L}$ through the U-Net encoder $ E_{img}$, where $B$ represents batch and $C$ represents channels. An image adapter $A_{img}$ is used to map every batch $B$ of raw high-level features to the embedding $H_v^a \in \mathbb{R}^{B*D}$.

\begin{align}
H_v &= E_{img}(x_{img}),  \\
H_v^a &= A_{img}(H_v) ,
\end{align}

To match the shape of $H_t$, we duplicate $H_v^a$ according to the class number K. We define:

\begin{align}
\text{rep}(H,k)=concat[\underbrace{H,H,\ldots,H}_{k\text{ times}}]\label{eq:rep},
\end{align}

then, we obtain the result $H_v^a \in \mathbb{R}^{B*K*D}$,

\begin{align}
H_v^a&= rep(A_{text}(H_v),K).
\end{align}

\subsection{Attention-based self adaptive learning  pipeline}
The vision language models have shown promising results across various tasks, attributable to their generalizability and interpretability. However, they often face the image and text distribution gap when applied to downstream tasks. For example, a medical segmentation dataset may have task-specific image styles and text formats that are not included in the pre-trained data sources. Therefore, how to fine-tune the pre-trained model at a lower cost and how to fuse different modalities of embeddings can be a noteworthy problem. We propose an attention-based self-adaptive learning pipeline to address the problem effectively.

In detail, the capability of CLIP is rooted in the natural image-text pairs. We enhance it for medical image segmentation tasks through fine-tuning. During training, the pre-trained CLIP encoder maintained frozen instead of fully adjusting all parameters to reduce the computing workload. We devise an adapter module and integrate it into designated positions shown in Fig~\ref{fig:method}. The adapter consists of a down-projection, ReLU activation, and up-projection with batch normalization and self-attention sequentially. The down-projection compresses the given embedding into a lower dimension using an MLP layer. At the same time, the up-projection expands the compressed embedding back to its original dimension using another MLP layer. Self-attention calculation captures the correlation of each class. The adapter trains CLIP embedding at a low cost with frozen parameters while intensely learning the attention of each class, guiding the segmentation model through the fusion method. Additionally, a trainable adapter is introduced into the vision model component, which is based on a pre-trained U-Net due to its manageable parameter size, making it an efficient starting point for training. This adapter facilitates the transition from image features to embeddings, enhancing the segmentation process.

In terms of fusing text and image embeddings after adopting the vision language model as the backbone,
Lin et al.~\cite{lin2023clip}, Zhou et al.~\cite{ zhou2023zegclip} and Liu et al.~\cite{liu2023clip} adopt simple strategies, such as direct plus or concatenation. 
Differently, we adopt the cross-attention(CA) module to integrate the two domain embeddings adaptively. The attention function serves as the operation to discover inner relationships from one modality to another. We have used the aforementioned adapters to get text embedding  $H_t^a \in \mathbb{R}^{B*K*D}$ and image embedding $H_v^a\in \mathbb{R}^{K*D}$, for every batch $H_t^a(b)$, we calculate the attention scores $H_f$:
\begin{equation}
f_{\text{CA}}(H) = \text{softmax}\left(\frac{q(H_t^a)^T k(H_t^a +  H_v^a(b))} {\sqrt{d_k}}\right) v(H_v^a(b)).
\end{equation}
The input sequences of these two modalities are identically ordered 
in our input. Based on this, contextual clues can be propagated between modalities. As for the cross-attention module's detail, we choose the text embedding to be query (Q), image embedding to be value (V), 
 and the plus of them to be the key (K). With language embedding's guidance, more precise features can be automatically selected rather than concatenated or plus in a hand-crafted way. The alignment of image and text embedding $H_f$ uses a multi-layer perceptron (MLP) to generate parameters ($\theta_k$). Three sequential convolutional layers with 1 × 1 × 1 kernels filling with ($\theta_k$) convert vision decoder output features $F$ into k  predictions, where $P_k = Sigmoid((F * \theta_{k_1}) * \theta_{k_2}) * \theta_{k_3}), \theta_k = \left\{\theta_{k_1}, \theta_{k_2}, \theta_{k_3}\right\}$. $*$ represents convolution operation. For each class $k$, we get every foreground class $P_k \in \mathbb{R}^{1\times H \times W \times L}$. After that, we merge k classes of prediction into one prediction $P$, shown in Fig~\ref{fig:method}. $P_k$ is supervised by label $Y_k$, where the overall loss is represented as:
\begin{equation}
    \mathcal{L}_{sup} = \frac{1}{|B|}\sum_{i=1}^{|B|} \left[ \mathcal{L}_S(P_k, Y_k)  \right], \label{eq:suploss}
\end{equation}
where $\mathcal{L}_S = \frac{1}{2} \left[ \mathcal{L}_{Dice} + \mathcal{L}_{ce} \right]$; $\mathcal{L}_{Dice}$ and $\mathcal{L}_{ce}$ represent the Dice and cross-entropy losses, respectively. The pseudo code of our framework can be found on Alg.~\ref{alg:training_procedure}

\begin{figure}[!htp]
    \centering
    \includegraphics[scale=0.39]{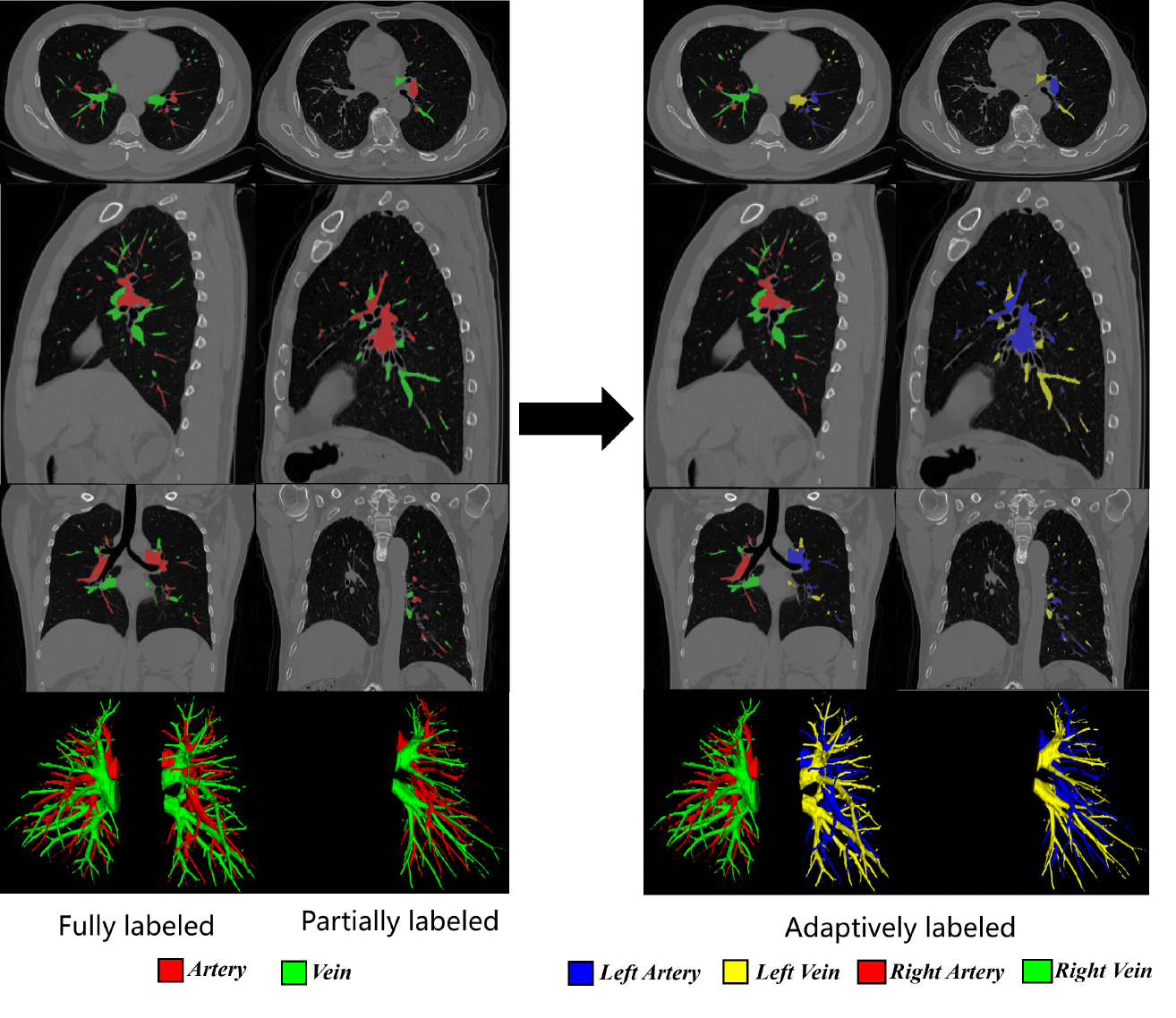}
    \caption{This pic  shows the adaptation  of our data. We only change the label of the artery/vein without making any other adjustments.
    }
    \label{fig:adaption_data}
\end{figure}

\begin{algorithm}[h]
\caption{Training Procedure of Our Method}
\label{alg:training_procedure}
\footnotesize
\SetAlgoLined

\KwData{\\
  Half-labeled data $D^p=\{(x_i, y_i)\}_{i=1}^{N}(y_i = 1,2)$, \\
  Fully labeled data $D^f = \{(x_i, y_i)\}_{i=1}^{M}(y_i = 1,2)$}
\KwIn{\\
Processed labeled data  $D^m = \{(x_i, y_i)\}_{i=1}^{N+M}(y_i = 1:4)$,
Segment network encoder $S_{enc}$, Segment network decoder  $S_{dec}$, CLIP encoder $T_{enc}$, 
batchsize  $\mathcal{B}$, number of classes $num\_class$, max epoch $E_{max}$}
\KwOut{\\
Trained weights of model $ f(\cdot; \theta)$ }

\For{epoch in $E_{max}$}{
    \For{batch in  $\mathcal{B}$}{
        Freeze CLIP model  $T_{enc}$\\
        Initialize  $num\_class$ Adapter $Adp_{text}$ \\
        Initialize one Adapter $Adp_{img}$ \\
        Get text embedding feature  $H_t \leftarrow T_{enc}$\\
        Get segment embedding feature $H_v \leftarrow S_{enc}(x_i)$\\
        Calculate $H_t^a \leftarrow Adp_{text}(t)$    \\ 
        Calculate  $H_v^a(batch) \leftarrow Adp_{img}(H_v^a(batch))$   \\
        $H_v^a\leftarrow rep(H_v^a,num_class)  
        $ according to Eqs.~\eqref{eq:rep}\\

        $H_f \leftarrow Cross\_attention(H_t^a ,H_v^a +H_t^a ,H_v^a )$ // get mixed feature map\\
         Calculate $\mathcal{L}_{sup}$ according to Eqs.~\eqref{eq:suploss}\\
        Update $\mathcal{L}_{sup}$ according to Eqs.~\eqref{eq:suploss}\\
    }
    epoch = epoch + 1
}

return model $f(\cdot; \theta)$
\end{algorithm}

\begin{table}[!htbp]

\centering
\caption{nnU-Net preprocess experiments under different  dataset setting.
\label{table:nnunet}
}
\begin{tabular}{ 
    m{1.7cm}<{\centering}  
    m{2.5cm}<{\centering}  
    m{1.7cm}<{\centering} 
}
\toprule[2pt]
\scriptsize
Methods & DSC(\%) $\uparrow$  & Jaccard(\%) $\uparrow$ \\ 
\midrule[1pt] 
\multicolumn{3}{c}{train:val:test = 64:6:7(Fully labeled)} \\
\midrule[1pt] 
nnU-Net & $66.14\pm{0.27}$  & $54.31\pm{0.15}$\\ 
Ours &$69.34\pm{0.24}$ & $58.32\pm{0.11}$ \\ 
\midrule[1pt] 
\multicolumn{3}{c}{train:val:test = 502:72:143(Fully labeled + Half-labeled)} \\ 
\midrule[1pt]  
nnU-Net &$67.19\pm{0.12}$ &$56.45\pm{0.34}$ \\ 
Ours &$76.22\pm{0.76}$ & $62.74\pm{0.45}$ \\ 
\bottomrule[2pt]
\end{tabular}
\end{table}


\section{Dataset and implementation detail}


\begin{figure*}[!htp]
\centering
\includegraphics[scale=0.64]{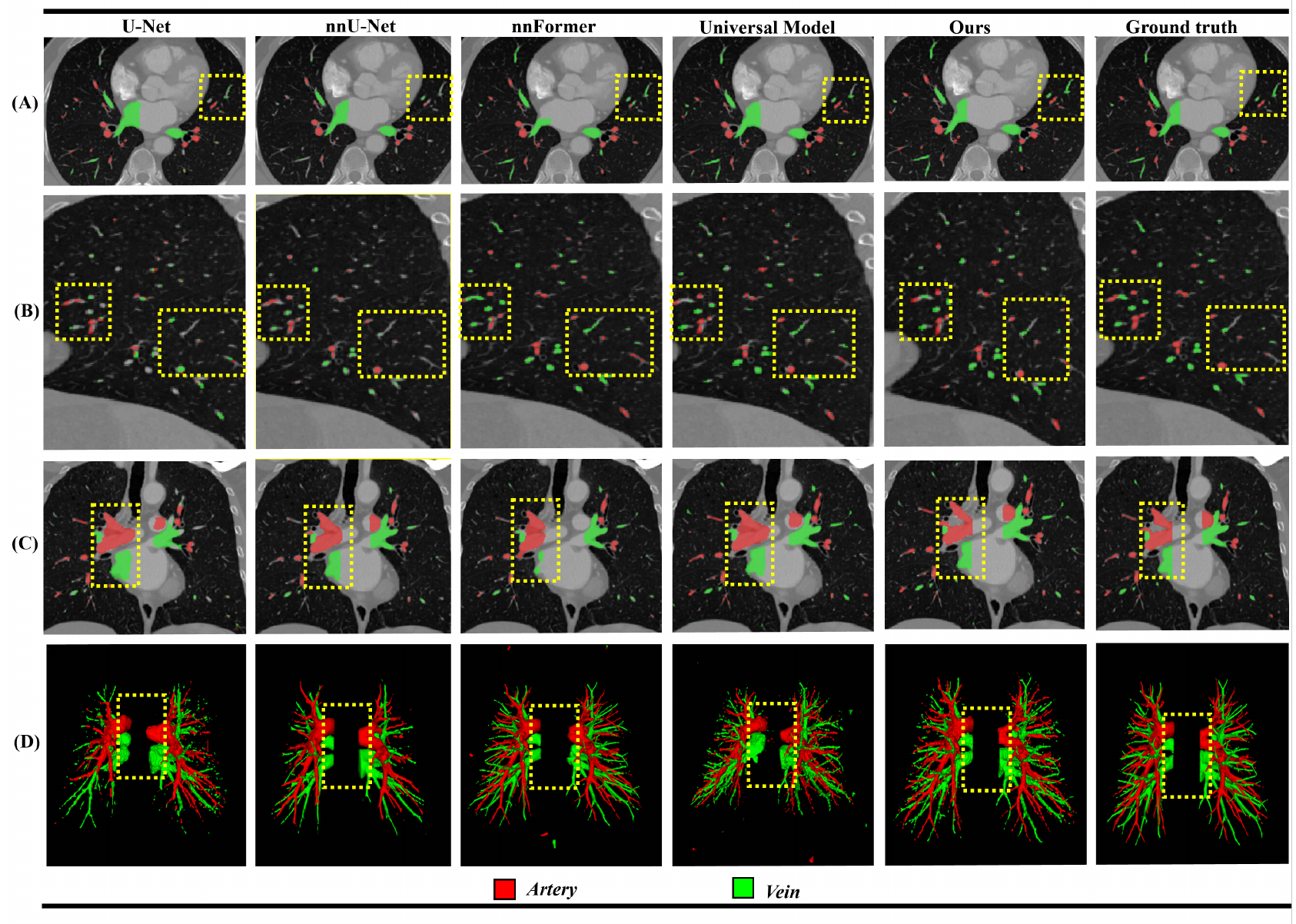}
\caption{Visualization of segmentation results on our dataset with zoomed-in views for enhanced clarity. (A-D) The segmentation results of one case are presented in the transverse section, coronal section, sagittal section, and 3D view, respectively. The regions enclosed by the dashed yellow boxes indicate misclassification executed by other models; our method can segment those regions closer to the ground truth. Mask colors for vessels have been standardized to red for arteries and green for veins to better visualize the differences between the methods.
\label{fig:comperation_results}
}
\end{figure*}

\subsection{\textbf{Dataset:}} We present a large-scale pulmonary vessel segmentation dataset collected from a real-world local hospital, comprising a total of 718 3D CT volumes provided in compressed NIFTI (.nii.gz) format. Among these, the pulmonary arteries and veins are manually annotated, where 79 CT scans are fully labeled and 639 CT scans are half-labeled, indicating the involvement of either the left lung or the right lung, as depicted in Fig.~\ref{fig:adaption_data}. In clinical practice, most patients typically exhibit disease in only one lung, with only a small proportion affected in both lungs. Therefore, it is reasonable to compile a dataset that combines fully labeled and half-labeled CT scans. The sizes of these CT volumes range from 512×512×169 to 512×512×985, with varying slice thicknesses from 0.62 to 1.25 mm. Annotations are obtained from five junior clinicians (with one to five years of experience) who used MIMICS to manually refine the segmentation results under the supervision of two board-certified radiologists. Finally, a senior radiologist with over ten years of experience verified and refined the annotations. We will make the dataset and annotations publicly available upon the acceptance of this work.
We divide the labeled data into training, validation, and test sets at a ratio of 7:1:2, where 502 volumes are designated as the Training Dataset; 72 volumes are allocated for validation and 143 volumes for testing. Results are shown on Tabel~\ref{table:comparison-methods}. All experiments are reported as $mean_{std}$  with three repeated trials in this work.


\subsection{\textbf{Implementation detail:}} Our model is implemented with U-Net as the backbone and optimizing the parameters via AdamW \cite{loshchilov2017decoupled}. The training utilizes a batch size of 4 and a patch size of 96 × 96 × 96. The default initial learning rate is set to 8e-4, with a momentum of 0.9 and a decay of 1e-5. The framework is implemented in MONAI version 0.9.05. The best model is selected within 200 epochs by evaluating the validation metrics. Models are trained on a single 40 GB NVIDIA A100 card. The Dice Similarity Coefficient (DSC), Normalized Surface Distance (NSD),  Jaccard and 95\% Hausdorff distance (HD95) are used to evaluate vessel segmentation performance in this work.

\textbf{Data augmentation:} 
As mentioned above, our data consists of both fully and half-labeled datasets with annotations: for example, we use 1 to represent the annotations of the pulmonary artery and 2 for the pulmonary vein. However, due to the complexity of tubular structures and the close intensity values of pulmonary vessels and airways in CT scans, segmenting the pulmonary vessels is quite challenging. Therefore, we adopt strong augmentation techniques for both CT input and label ground truth.

\textbf{CT augmentation: }
Firstly, we adjusted the window width and level to meet the appropriate Hounsfield unit for pulmonary vessels, ranging from -700 to 300. Then, we calculate the Hessian matrix of the CT and obtain the eigenvalue to fill the Z-axis, which strengthens the CT's tubular structures, as shown in Fig~\ref{fig:method}.

\textbf{Label augmentation: }  We convert the three-class segmentation task (\textit{i.e.} background, artery, vein) into a five-class segmentation task. For example, we represent 1 for the left pulmonary artery, 2 for the left pulmonary vein, 3 for the right pulmonary artery, 4 for the right pulmonary vein and 0 for the background. This method can improve the model's performance significantly, not only for us but other state-of-the-art methods as well. It can be seen from Table~\ref{table:nnunet} that when adding the half-labeled data into training without adaptive augmentation, the DSC slightly increased from 66.14\% to 67.19\%. This means that the incomplete label is not exploited in this setting. Instead, our method utilizes these settings and gets the full potential of the half-labeled data. Therefore, we choose a simple but efficient augmentation for our incomplete label.

\begin{table}[!htbp]
\centering
\caption{Ablation study of every componet of our framework.UM indicates Univer Model,DA indicates Data Augmentation,AAP indicates Attention-based self Adaptive learning Pipeline }
\label{table:ablation_method}
\begin{tabular}{llccccc}
\toprule[2pt]
UM & DA & AAP  & DSC(\%) $\uparrow$ & Jaccard(\%)$\uparrow$ & NSD$\downarrow$ & HD95$\downarrow$\\
\midrule[1pt]
\checkmark & & &$64.49_{0.45}$ & $56.32_{0.15}$ &$0.98_{0.03}$ &$47.34_{0.35}$  \\
\checkmark & \checkmark & & $68.38_{0.25}$ & $58.33_{0.02}$ &$0.91_{0.06}$ &$46.43_{0.01}$ \\
\checkmark & \checkmark & \checkmark & $76.22_{0.76}$ & $62.74_{0.24}$ &$0.86_{0.43}$ &$14.48_{0.22}$\\
\bottomrule[2pt]
\end{tabular}
\end{table}



\begin{table*}[!htbp]
\centering
\caption{Summary of experimental results on the proposed dataset, detailing the comparison of our methods with existing approaches for the classes of artery and vein. The reported metrics include DSC, Jaccard, NSD, and HD95. Metrics are presented in the form of $mean \pm std$, where each method is evaluated over three trials for averaging.}
\label{table:comparison-methods}
\begin{tabular}{lcccccccccc}
\toprule[2pt]
\multirow{2}{*}{Methods} & \multicolumn{4}{c}{Artery} & \phantom{abc}& \multicolumn{4}{c}{Vein} & \multirow{2}{*}{Mean DSC} \\
\cmidrule{2-5} \cmidrule{7-10}
& DSC(\%) $\uparrow$ & Jaccard(\%)$\uparrow$ & NSD$\downarrow$ & HD95$\downarrow$ && DSC(\%)$\uparrow$ & Jaccard(\%)$\uparrow$ & NSD$\downarrow$ & HD95 $\downarrow$\\

\midrule[1pt]
U-Net \cite{ronneberger2015u}& $\text{61.23}_{0.48}$ & $\text{48.34}_{0.38}$ & $\text{1.94}_{0.19}$ & $\text{132.23}_{1.32}$ && $\text{59.38}_{0.49}$ & $\text{49.32}_{0.49}$ & $\text{1.43}_{0.14}$ & $\text{110.76}_{1.11}$ & $\text{60.49}_{0.60}$\\
nnU-Net \cite{isensee2021nnunet} & $\text{67.52}_{0.63}$ & $\text{56.45}_{0.56}$ & $\text{0.81}_{0.05}$ & $\text{43.76}_{0.51}$ && $\text{66.86}_{0.56}$ & $\text{55.81}_{0.21}$ & $\text{0.77}_{0.08}$ & $\text{44.35}_{0.44}$ & $\text{67.19}_{0.67}$ \\
nnFormer \cite{zhou2021nnformer} & $\text{64.46}_{0.51}$ & $\text{51.32}_{0.12}$ & $\text{0.89}_{0.06}$ & $\text{86.61}_{0.87}$ && $\text{63.25}_{0.82}$ & $\text{52.67}_{0.53}$ & $\text{0.85}_{0.09}$ & $\text{96.32}_{0.96}$ & $\text{63.85}_{0.64}$ \\
Universal Model \cite{liu2023clip} & $\text{70.24}_{0.57}$ & $\text{57.25}_{0.57}$ & $\text{0.79}_{0.02}$ & $\text{25.34}_{1.78}$ && $\text{69.32}_{0.54}$ & $\text{54.32}_{0.55}$ & $\text{0.77}_{0.06}$ & $\text{28.97}_{1.16}$ & $\text{69.78}_{0.69}$\\
Ours & $\text{77.26}_{0.64}$ & $\text{64.13}_{0.61}$ & $\text{0.86}_{0.05}$ & $\text{13.72}_{0.14}$ && $\text{75.19}_{0.23}$ & $\text{61.34}_{0.15}$ & $\text{0.86}_{0.09}$ & $\text{15.24}_{0.15}$ & $\text{76.22}_{0.76}$ \\
\bottomrule[2pt]
\end{tabular}
\end{table*}


\section{Experiments}
\subsection{Ablation studies}
We conduct ablation studies to evaluate every component of our proposed pipeline. 
The quantitative results of the different methods are presented in Table.~\ref{table:ablation_method} with DSC, Jaccard, NSD and HD95. The pre-trained Univer Model(UM)~\cite{liu2023clip} is used as our baseline. We first use a simple Data Augmentation(DM) to effectively use our half-labeled data, contributing 
a performance gain of over 3.89\% DSC, 2.01\% Jaccard over baseline, reduces 0.007 NSD and 0.91 HD95 compared with baseline. Subsequently, our proposed attention-based self-adaptive learning pipeline is introduced to fine-tune the pre-trained model and align text representations with image representations with an adaptive attention mechanism. We observe a further increment of 7.84\% in DSC, 4.41\% in Jaccard and a significant decrement of  31.95 in HD95, 0.05 in NSD. 

\subsection{Comparison of quantitative results on test dataset }

Table \ref{table:comparison-methods} presents a qualitative comparison of our proposed vessel test dataset, which consists of 143 CT volumes, against other state-of-the-art methods. As shown in Table \ref{table:nnunet}, nnU-Net~\cite{isensee2021nnunet} demonstrated promising results, achieving a mean DSC of 66.14\% and a mean Jaccard of 54.31\% when trained on a fully labeled dataset. However, when a substantial amount of half-labeled data is incorporated into the training process, nnU-Net only improved by 1.05\% in DSC. In contrast, our framework achieved a DSC of 69.34\% and a Jaccard of 58.32\% with a small fully labeled dataset, and demonstrated a significant increase of 6.88\% in DSC when utilizing all available labeled data. We speculate that the notable decrease in performance of nnU-Net compared to our framework is attributable to the ambiguity introduced by the additional half-labeled data. For instance, fully labeled CT scans annotate both left and right arteries, whereas half-labeled scans may only specify left or right arteries, leaving the other unlabeled. Thus, merely adding more data does not necessarily enhance model training. Therefore, we report the performance metrics of other state-of-the-art methods trained on the fully labeled dataset but tested on the 143 CT volumes. The universal model achieved a mean DSC of 69.78\%, as shown in Table \ref{table:comparison-methods}, which is 5.29\% higher than the universal model trained on all dataset settings, as depicted in Table \ref{table:ablation_method}. Training with a much smaller dataset while attaining a significantly higher mean DSC supports our hypothesis. Compared to the vanilla 3D U-Net, all other methods outperformed it in terms of DSC and Jaccard, although they demonstrated smaller DSC and HD95 scores for both artery and vein. Our method achieved a DSC of 77.26\%, a Jaccard of 64.13\%, a NSD of 0.86, and an HD95 of 13.72 for arteries, outperforming the baseline universal model~\cite{liu2023clip} by 7.02\% in DSC and 5.93\% in Jaccard, as well as surpassing the supervised self-configuring model nnU-Net by 9.74\% in DSC and 7.68\% in Jaccard. Additionally, it outperformed the attention-based self-configuring model nnFormer by 12.8\% in DSC and 12.81\% in Jaccard. Overall, our model significantly surpasses all other compared methods, achieving superior state-of-the-art performance. Furthermore, Figure \ref{fig:comperation_results} displays the visualization of segmentation results for our method and others, illustrating that our method segments both veins and arteries closer to the ground truth.
\section{Conclusion}

This work introduces a novel segmentation framework, integrating language-vision models with a self-adaptive feature learning pipeline and a designated data augmentation strategy. We leverage our partially annotated dataset to adhere to the best practices from large vision-language models. The framework incorporates our proposed adapter for fine-tuning CLIP embeddings, enhanced with self-attention to capture inter-class relationships. Furthermore, a cross-attention mechanism is 
seamlessly integrated to promote the effective fusion of the vision model with the segmentation model. We present the most extensive clinical dataset to date for pulmonary artery vein segmentation, comprising 718 high-quality CT volumes. Empirical evaluation demonstrates that our  framework sets a new benchmark on this dataset, achieving an average DSC of 76.22\% on a test set of 143 volumes, surpassing nnU-Net by over 10\%. These results affirm the superiority of our framework in the challenging task of pulmonary vessel segmentation against current state-of-the-art methods.

\bibliographystyle{IEEEbib}
\bibliography{icme2025references}

\vspace{12pt}

\end{document}